\documentclass[runningheads]{llncs}
\usepackage[T1]{fontenc}
\usepackage{graphicx}
\usepackage{silence}
\usepackage{amsmath}
\usepackage{amssymb}
\usepackage{booktabs,multirow}
\begin{document}
\title{Unsupervised Domain Adaptation for Symbol Spotting in Historical Encrypted Manuscripts}

%
\titlerunning{Unsupervised Domain Adaptation for Symbol Spotting}
%
\author{
Giuseppe De Gregorio\inst{1}\orcidID{0000-0002-8195-4118} \and
Alicia Fornés\inst{1}\orcidID{0000-0002-9692-5336} \and
Lei Kang\inst{1}\orcidID{0000-0002-1962-3916}
Beáta Megyesi\inst{2}\orcidID{0000-0002-4838-6518}
}

\authorrunning{G. De Gregorio et al.}
%
\institute{Computer Vision Center (CVC), Universitat Autònoma de Barcelona, Spain 
\email{\{gdegregorio,afornes,lkang\}@cvc.uab.cat}
\and
 Stockholm University, Sweden \\
\email{beata.megyesi@ling.su.se}
}
\maketitle              
\begin{abstract}
The decipherment of historical encrypted manuscripts poses a fundamental challenge in Digital Humanities: before any transcription can begin, the symbol inventory of the underlying cipher alphabet must first be identified and characterized. We address this challenge through symbol spotting: given a candidate alphabet specified as a set of rendered font glyphs, the task is to determine whether and where its characters appear in an unseen handwritten document, without any labeled examples from the target script. The main difficulty lies in the domain gap between clean, digitally rendered font queries and degraded handwritten manuscript symbols. We propose a three-stage pipeline that bridges this gap without manual annotation, combining a joint SimCLR+DANN encoder for domain-invariant glyph representations with an embedding-space style-adaptation mechanism applied at retrieval time, requiring no re-training. 
Experiments on fourteen pages from seven encrypted manuscript collections show that our method outperforms zero-shot foundation models, including CLIP and DINOv2, by a large margin ($+0.194$ P@1 over CLIP ViT-L/14), and surpasses task-specific trained baselines by $+0.138$ P@1. 
We further demonstrate that the Raw-Cover metric, computed in a fully unsupervised setting, provides a meaningful script-family fingerprint that identifies the underlying alphabet of an unknown document. This capability is of direct practical relevance to palaeographers, historians, and other researchers working with undeciphered manuscripts.
\begingroup
\let\thefootnote\relax\footnote{
The code is publicly available at \url{https://github.com/giuseppedeg/symbol-spotting-encrypted-manuscripts}}
\endgroup.

\end{abstract}

\section{Introduction}
\label{sec:intro}

Recent progress in handwritten text recognition has substantially improved access to historical manuscripts written in known scripts and languages, particularly when sufficient annotated training data is available. In contrast, documents written in non-standard, rare, undeciphered, or otherwise unknown scripts remain largely beyond the reach of standard computational approaches, since both labelled data and prior knowledge of the underlying symbol system may be limited or absent.
The DECODE database~\cite{DECODE2022,Megyesi2026LT4HALA} has archived hundreds of such manuscripts, spanning cipher alphabets, personal notations, and invented writing systems whose interpretation remains unresolved. These sources raise methodological questions that differ from those addressed by conventional recognition pipelines, as they require methods capable of operating under conditions of extreme data scarcity and uncertain script structure.

A fundamental prerequisite for deciphering such documents is symbol spotting: given a set of candidate symbols, determining whether and where they appear in the manuscript. In the simplest case, this is a same-domain problem. The candidate symbols and the manuscript symbols share the same visual characteristics, and the task reduces to template matching or learned similarity. However, the definition of the candidate symbol set is itself a non-trivial challenge when the underlying script is unknown or only partially understood. A natural strategy is to query the manuscript against known alphabets (Greek, Latin, Runic, or other historically attested scripts) and use the retrieval signal to identify which character families are present. The most readily available source of such candidate symbols is the digital Unicode standard, for which high-quality fonts exist that can render any registered character as a clean glyph image. This approach is practical and requires no manual annotation, but it introduces a fundamental asymmetry: the query symbols are clean, regularly spaced, digitally rendered glyphs, while the gallery symbols are extracted from degraded, handwritten pages with variable ink, noise, and stroke irregularities. The problem therefore becomes inherently cross-domain, and it is this domain gap between font-rendered queries and manuscript-extracted gallery symbols that is the central challenge we address.

Existing approaches to word and symbol spotting in historical documents broadly fall into two categories. Training-based methods~\cite{SouibguiFKT20,JemniASKC26,Wolf2020} achieve strong results but require labelled examples from the target manuscript, a condition rarely met for encrypted scripts. Zero-shot methods based on large vision-language models such as CLIP~\cite{RadfordKHRGASAM21} or self-supervised transformers such as DINOv2~\cite{OquabDMVSKFHMEA24} offer an appealing alternative, but as we demonstrate empirically, their patch-based architectures are poorly suited to the retrieval of isolated glyphs at the small resolutions typical of symbol crops.
We propose a three-stage pipeline that bridges this gap without requiring any labelled manuscript data. First, individual glyphs are extracted from manuscript pages, and candidate alphabet symbols are rendered from fonts that approximate historical handwriting styles. Second, a joint DANN+SimCLR encoder is trained to produce domain-invariant embeddings: a contrastive objective~\cite{ChenK0H20} encourages structurally similar symbols to cluster together, while a domain adversarial network with gradient reversal~\cite{Ganin2016} penalises features that betray their domain of origin. Third, at retrieval time, a lightweight embedding-space style adaptation shifts each font query towards the local style manifold of the target document, further closing the residual domain gap without any re-training.
Our contributions are as follows:

\begin{itemize}
    \item We frame zero-shot symbol spotting in encrypted manuscripts as a cross-domain retrieval problem and propose a unified pipeline addressing all three stages: extraction, representation, and retrieval.
    \item We introduce an embedding-space style adaptation mechanism that operates on-the-fly at inference time and consistently improves retrieval performance across all tested configurations.
    \item We provide an extensive empirical evaluation on fourteen pages from seven encrypted manuscript collections, comparing against zero-shot baselines and task-specific trained models, and conduct a thorough ablation study isolating the contribution of each component.
    \item We demonstrate that Raw-Cover@k, computed in a fully unsupervised setting, can serve as a script signature for identifying the underlying alphabet family of an unknown document.
\end{itemize}

\section{Related Work}
\label{sec:sota}

\paragraph{Supervised Symbol and Word Spotting.}
The dominant paradigm in historical document analysis frames spotting as a supervised retrieval problem: a model is trained on labelled examples from the target collection and then used to rank candidate images by similarity to a query. 
Early approaches relied on handcrafted descriptors such as HOG features, projection profiles, and DTW-based alignment, establishing the first retrieval benchmarks on collections like George Washington~\cite{RathM07a}.
The advent of deep learning shifted the paradigm decisively: convolutional and recurrent networks trained with PHOC embeddings~\cite{SudholtF16} and Siamese networks with contrastive losses~\cite{Fathallah2023} have since set strong benchmarks on well-annotated collections such as Bentham.
For the specific sub-problem of cipher recognition, Souibgui et al.~\cite{SouibguiFKT20} propose a few-shot symbol 
spotting and recognition pipeline: symbols are first detected in line images, then decoded by mapping similarity scores to the transcribed sequence, with training on synthetic data and optional fine-tuning on a handful of labelled pages. While effective, these methods share a common requirement: at least some annotated material from the target script. For encrypted manuscripts where the alphabet itself may be unknown, this assumption cannot be satisfied.

\paragraph{Weakly Supervised and Self-Supervised Approaches.}
To relax the annotation bottleneck, a growing body of work exploits weak supervision or self-supervised pretraining. Wolf and Fink~\cite{Wolf2020} demonstrate that annotation-free word spotting is achievable by combining a lexicon-guided pseudo-labelling scheme with self-supervised feature learning, attaining query-by-string performance without any ground-truth transcriptions. Jemni et al.~\cite{JemniASKC26} propose ST-KeyS, a masked autoencoder pretraining strategy for keyword spotting based on vision transformers, which outperforms supervised baselines on several benchmarks when fine-tuned with minimal annotations. In the broader computer vision literature, SimCLR~\cite{ChenK0H20} established that strong visual representations can be learned purely from unlabelled data via contrastive augmentation pairs, a principle we adopt and extend with domain-adversarial training. Domain-Adversarial Neural Networks~\cite{Ganin2016} provide the complementary mechanism: by routing encoder features through a Gradient Reversal Layer, the model is penalised for any representation that betrays its domain of origin, producing embeddings that generalise across the font-to-manuscript gap.

\paragraph{Zero-Shot Foundation Models.}
The recent emergence of large-scale vision foundation models has renewed interest in zero-shot retrieval, where no task-specific training is performed at all. CLIP~\cite{RadfordKHRGASAM21} aligns visual and textual representations through contrastive pretraining on billions of image-text pairs, enabling strong zero-shot classification via natural language queries. DINOv2~\cite{OquabDMVSKFHMEA24} takes a purely visual route, training a Vision Transformer on 142M curated images with a self-supervised objective, producing features that transfer remarkably well across diverse downstream tasks without fine-tuning. Despite their impressive general-purpose performance, we find empirically that both families of models underperform task-specific approaches on isolated symbol retrieval: CLIP's image-text alignment is ill-suited to a purely visual similarity task, while DINOv2's patch-based ViT architecture loses discriminative power when processing the small glyph crops typical of symbol spotting pipelines. This observation motivates our domain-specific training strategy and positions the present work as complementary to, rather than competing with, foundation model approaches.

\section{Method}
\label{sec:method}

\begin{figure}[t]
  \centering
  \includegraphics[width=\columnwidth]{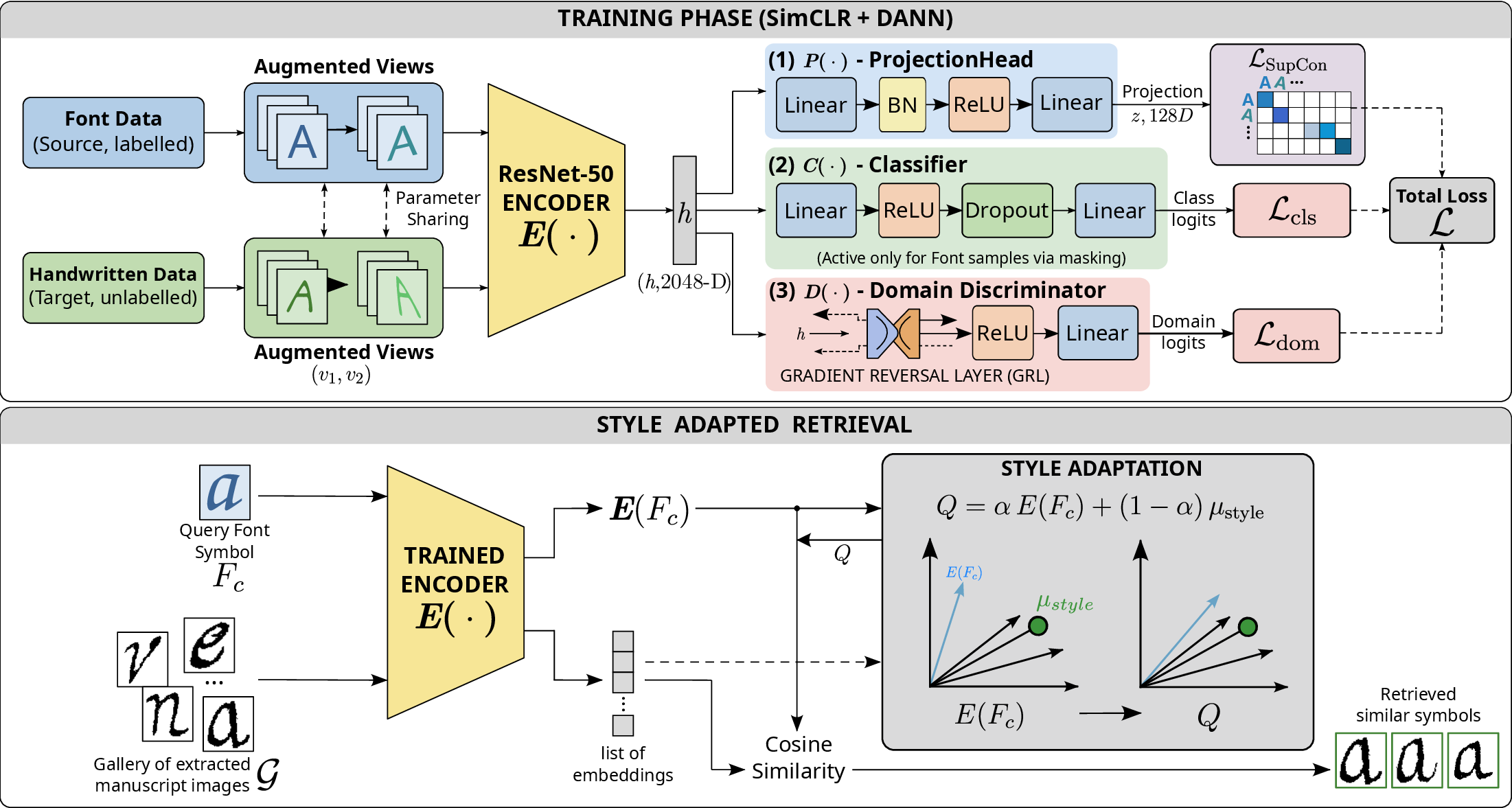}
  \caption{Overview of the proposed system architecture. (Top) The joint training framework utilizes SimCLR for structural feature learning and a Domain Adversarial Neural Network (DANN) to achieve font-to-manuscript invariance. (Bottom) Schematic representation of the style-adapted retrieval process, where font queries are dynamically adjusted toward the local document manifold via local centroid interpolation.}
  \label{fig:architecture}
\end{figure}

Our pipeline (Figure~\ref{fig:architecture}) addresses a zero-shot symbol-spotting task: given a historical handwritten document whose script is unknown, determine whether the characters of a candidate alphabet appear in it. The alphabet is specified only as a set of rendered font glyphs; no labelled examples from the manuscript are available.

\noindent The approach consists of three stages:
I) symbol extraction and rendering; II) domain-adaptive representation learning; III) embedding-space style adaptation at retrieval time.

\subsection{Symbol Extraction and Rendering}
\label{sec:extraction}

\paragraph{Manuscript symbols.}
Historical documents are first binarized using the Sauvola local thresholding algorithm~\cite{sauvola2000}, which adapts the local threshold to the background luminance and is therefore robust to uneven illumination and parchment degradation. Because the target manuscripts are written in a \emph{detached} style (all symbols are well separated with minimal ligatures), individual glyphs can be reliably isolated by analysing the connected components (CC) of the foreground ink pixels.
Each CC bounding box is cropped and padded to produce a square image with a white background and black ink, yielding a large unlabelled gallery $\mathcal{G} = \{g_i\}_{i=1}^{N}$ of manuscript symbols.

\paragraph{Alphabet rendering.}
Candidate alphabets are provided as lists of Unicode code points. For each character $c$ in alphabet $\mathcal{A}$, we render a set of images $\{F_c^{(j)}\}_{j=1}^{M}$ using a collection of fonts that visually approximate historical handwriting styles. The resulting images share the same white-background, black-ink normalisation as the manuscript crops, reducing the domain gap at the pixel level. A non-trivial gap nonetheless persists at the texture and stroke-regularity levels, which the subsequent training stage is designed to bridge.

\subsection{Domain-Adaptive Representation Learning}
\label{sec:training}

To build a shared embedding space for font and manuscript symbols, we train a network that jointly optimises a supervised contrastive objective on labelled font glyphs and a domain adversarial objective across both domains. The goal is an encoder whose representations are discriminative with respect to character identity yet invariant to the font-versus-manuscript domain shift.

\paragraph{Encoder architecture.}
We adopt a ResNet-50~\cite{he2016} backbone $E(\cdot)$, with the final classification head replaced by an identity mapping, that produces a 2048-D feature vector $\mathbf{h} = E(x)$.  A two-layer MLP projection head $P(\cdot)$ maps $\mathbf{h}$ to a 128-D space $\mathbf{z} = P(\mathbf{h})$ used exclusively for the contrastive loss. A separate two-layer MLP classifier $C(\cdot)$ and a two-layer domain discriminator $D(\cdot)$ are both attached directly to $h$.

\paragraph{Supervised contrastive loss.}
Because manuscript symbols carry no class labels, we cannot define inter-sample positives for them. For font samples, however, class labels are available and should be fully exploited. We therefore adopt the Supervised Contrastive loss (SupCon)~\cite{KhoslaTWSTIMLK20}, which generalises NT-Xent~\cite{ChenK0H20} by treating all same-class samples in the batch as positives rather than only the augmentation pair. Given a mini-batch containing both font samples (with labels) and manuscript samples (without labels), we define the positive set $\mathcal{P}(i)$ for anchor $i$ as:
\begin{equation}
\mathcal{P}(i) =
\begin{cases}
\{j \neq i : y_j = y_i,\; d_j = 0\}, & \text{if } d_i = 0 \text{ (font)}, \\
\{i^+\}, & \text{if } d_i = 1 \text{ (manuscript)}.
\end{cases}
\label{eq:positive_set}
\end{equation}
where $d_i \in \{0, 1\}$ is the domain label and $i^+$ denotes the paired augmented view. Font anchors thus benefit from all same-class font samples in the batch as positives; manuscript anchors fall back to the standard augmentation pair. All samples, regardless of domain, contribute to the contrastive denominator as negatives. The loss is:
\begin{equation}
\mathcal{L}_{\mathrm{SupCon}} = -\frac{1}{|\mathcal{B}|}
\sum_{i \in \mathcal{B}} \frac{1}{|\mathcal{P}(i)|}
\sum_{p \in \mathcal{P}(i)}
\log \frac{\exp(\mathbf{z}_i \cdot \mathbf{z}_p / \tau)}
{\sum_{k \neq i} \exp(\mathbf{z}_i \cdot \mathbf{z}_k / \tau)},
\label{eq:supcon}
\end{equation}
where $\tau$ is the temperature and $\mathcal{B}$ is the full batch of $2N$ augmented views.

\paragraph{Supervised classification loss.}
A two-layer MLP classifier $C(\cdot)$ is attached to $h$ and trained with label-smoothed cross-entropy exclusively on font samples ($d{=}0$)):
\begin{equation}
\mathcal{L}_{\mathrm{cls}} =
  -\sum_{c} \tilde{y}_c \log C(\mathbf{h})_c,
\label{eq:cls}
\end{equation}
\noindent where $\tilde{y}$ is the label-smoothed target.

\paragraph{Domain adversarial loss (DANN).}
A domain discriminator $D(\cdot)$ attempts to distinguish font embeddings ($d{=}0$) from manuscript embeddings ($d{=}1$). Its input is routed through a Gradient Reversal Layer (GRL)~\cite{Ganin2016}, which negates the gradient during backpropagation and thereby forces the encoder to produce domain-invariant features:
\begin{equation}
\mathcal{L}_{\mathrm{dom}} =
  -\sum_{i} \bigl[d_i \log D(\tilde{\mathbf{h}}_i)
             + (1-d_i) \log (1-D(\tilde{\mathbf{h}}_i))\bigr],
\label{eq:dann}
\end{equation}
\noindent where $\tilde{\mathbf{h}}_i = \mathrm{GRL}_\lambda(\mathbf{h}_i)$.
The reversal coefficient is annealed as $\lambda = \frac{2}{1+e^{-10p}}-1$, with $p \in [0,1]$ denoting the fraction of completed training epochs.

\paragraph{Total objective.}
The three losses are combined as:
\begin{equation}
\mathcal{L} = \lambda_s \,\mathcal{L}_{\text{SupCon}} 
              + \mathcal{L}_{\text{cls}} 
              + \mathcal{L}_{\text{dom}}
\label{eq:total}
\end{equation}
\noindent where $\lambda_s \leq 1$ downweights the contrastive term relative to the classification signal, preventing the SupCon gradient from overwhelming the encoder during early training.

\paragraph{Training details.}
To guarantee that each mini-batch contains multiple same-class 
font samples, we use a class-balanced sampler that draws $K$ font classes per batch with $S$ samples each, where $K = \lfloor 2B / 3S \rfloor$ is derived from the target batch size $B$. This sampling strategy extends the class-balanced approach of~\cite{KhoslaTWSTIMLK20} to a mixed-domain setting, where unlabelled manuscript symbols are systematically included as hard negatives for both the contrastive and domain adversarial objectives. This is a design choice that proves critical for preventing SupCon from degenerating to self-supervised NT-Xent in the presence of unlabelled data. 

Training proceeds in two stages: for the first $T_w$ epochs the domain adversarial loss is disabled, allowing the encoder to build a stable feature space before the GRL begins to align the two domains. 
The model is trained with early stopping based on a composite score $\mathcal{L}_{\text{cls}} + |\ln 2 - \mathcal{L}_{\text{dom}}|$, which jointly minimises classification error and drives the domain discriminator towards its theoretical chance-level entropy of $\ln 2 \approx 0.693$.

\subsection{Retrieval with Embedding-Space Style Adaptation}
\label{sec:retrieval}
Despite DANN training, a residual domain shift persists at inference time: a single font glyph cannot capture the full distribution of a scribe's handwriting style, and the encoder's domain-invariant features may still place the font query at a slight offset from the manuscript cluster it belongs to. We address this with a lightweight style adaptation step that displaces each query towards the local style manifold of the target document at inference time, with no re-training required.

\paragraph{Style adaptation.}
Given a manuscript gallery $\mathcal{G}$ and a query font glyph $F_c$ for character $c$, we first compute the raw query embedding 
$E(F_c) \in \mathbb{R}^{d}$ and retrieve its $k$ nearest neighbours in $\mathcal{G}$ by cosine similarity. These neighbours approximate the local appearance of symbols in the target document that are visually close to $c$. Their centroid:
\begin{equation}
    \mu_{\text{style}} = \frac{1}{k} \sum_{s \in \mathcal{N}_k(F_c)} E(s)
    \label{eq:mu_style}
\end{equation}
summarises the average stroke style in that neighbourhood. The adapted query is then formed as a convex combination:
\begin{equation}
    Q = \alpha\, E(F_c) + (1 - \alpha)\, \mu_{\text{style}}
    \label{eq:adaptation}
\end{equation}
where $\alpha \in [0, 1]$ controls the trade-off between preserving character identity ($\alpha \to 1$) and conforming to the document style ($\alpha \to 0$). Intuitively, Eq.~\eqref{eq:adaptation} can be read as a local distribution estimate: it shifts the font query into the region of embedding space actually occupied by the target manuscript, without losing the discriminative identity of $c$. The adapted embedding $Q$ is $\ell_2$-normalised before the final retrieval step.

\paragraph{Retrieval.}
Final retrieval is performed by ranking all gallery embeddings by 
cosine similarity to $Q$. A match is accepted if the similarity exceeds a threshold $\tau_{\text{ret}}$, whose optimal value is determined empirically via the sensitivity analysis described in Section~\ref{sec:threshold}. For each query we retain at most $k_{\text{ret}} = 25$ results. In all experiments we use $k = 50$ neighbours for the style estimation and $\alpha = 0.7$, a value that consistently outperforms both extremes ($\alpha = 0$ and $\alpha = 1$) across all tested documents.

\section{Results}

\subsection{Datasets and Experimental Setup}
\label{sec:setup}
\paragraph{Dataset.}
We evaluate our method on fourteen pages drawn from seven collections of encrypted historical manuscripts, all sourced from the DECODE dataset~\cite{DECODE2022}. 
The collections span a wide range of cipher styles, scribal hands, and historical periods, providing a challenging and diverse testbed. All manuscripts are written in a detached script style, where individual symbols are well separated, a prerequisite for the connected-component extraction described in Section~\ref{sec:extraction}.
We evaluate on two pages per collection, identified in DECODE by the following record-ID/image-ID pairs: ASV (19/112, 19/114), Copiale (2/55 pp.~102, 2/55 pp.~103), Borg (210/1662, 210/1663), Maximilian (1471/6694, 1471/6698), Ramanacoil (1564/6996, 1564/6998), Runicus (Codex Runicus pp.~30,~97), and Vaticans (47/249, 47/250).

\begin{figure}[htbp]
    \centering
    \includegraphics[width=1\textwidth]{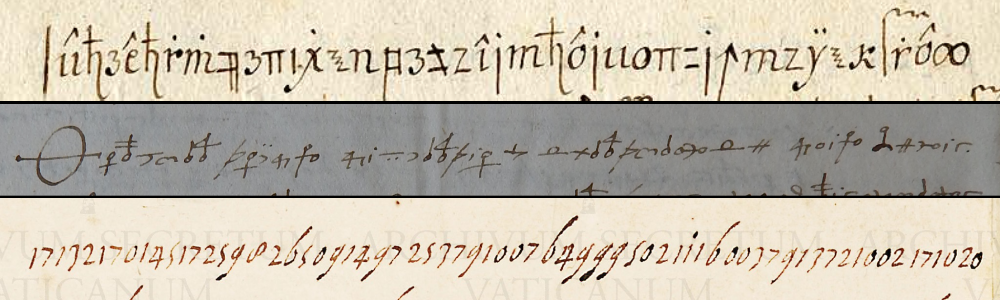}
    \caption{Three-line examples from (from top) the Copiale, Maximilian, and Vaticans collections.}
    \label{fig:data_examples}
\end{figure}

\paragraph{Alphabet construction.}
For each collection, a reference alphabet was constructed by filtering a set of known script families (Arabic, Digits, Latin, Greek, Zodiac, Esoteric, and Runic) against the symbols present in the documents of the collection. Concretely, for each script family we retained only those Unicode code points that appear at least once in the manuscript, ensuring that the query set reflects the actual symbolic vocabulary of the document rather than the full extent of the script. Characters were then rendered using the font pipeline described in Section~\ref{sec:extraction}. Additionally, a \emph{Copiale} alphabet was included as a special case: rather than being rendered from fonts, it was constructed by cropping individual character images directly from a manuscript page of the Copiale collection, providing a same-domain query baseline. A \emph{Cuneiform} alphabet was included as a negative control: none of its symbols appear in any of the test documents, so any retrieval hits for this alphabet can be attributed entirely to false positives.

\paragraph{Evaluation protocol.}
For each (document, alphabet, query character) triple, we retrieve up to $k_{\text{ret}} = 25$ gallery symbols ranked by cosine similarity. We evaluate performance using four complementary metrics.

\vspace{2mm}
\noindent \textbf{Precision@}$k$ measures the accuracy of the system within the top-$k$ retrieved results. Given a query, the system returns a ranked list of gallery symbols; we truncate this list at rank $k$ and compute:
\begin{equation}
    P@k = \frac{\text{correct matches in top-}k}{k}
    \label{eq:precision}
\end{equation}
A high Precision@$k$ indicates that the top results presented to the user are reliable, minimising the need to sift through noise.

\vspace{2mm}
\noindent \textbf{Cover@}$k$ measures what fraction of the target alphabet has been successfully retrieved within the top-$k$ results across all queries:
\begin{equation}
    \text{Cover@}k = 
    \frac{\text{unique correct symbols in top-}k}{\text{alphabet size}}
    \label{eq:cover}
\end{equation}
A value close to~1.0 indicates that the system has discovered almost the entire character set of the manuscript within a limited search budget. This metric is particularly relevant for paleographers, who are interested in identifying \emph{which} symbols are present in a document rather than just confirming the presence of known ones.

\vspace{2mm}
\noindent \textbf{Raw-Cover@}$k$ extends Cover@$k$ by counting both correct and incorrect retrievals, regardless of their accuracy:
\begin{equation}
    \text{Raw-Cover@}k = 
    \frac{\text{unique symbols (correct + incorrect) in top-}k}
         {\text{alphabet size}}
    \label{eq:raw_cover}
\end{equation}
Comparing Raw-Cover@$k$ with Cover@$k$ quantifies the system's hallucination rate: a large gap between the two indicates that the system is attempting broad coverage but retrieving many false positives. This metric is particularly useful in the unsupervised script identification scenario of Section~\ref{sec:script_id}, where no ground truth is available and Raw~Cover@$k$ can be computed without any labelled data.

\vspace{2mm}
\noindent \textbf{MRR} (Mean Reciprocal Rank) evaluates ranking efficiency by 
focusing on the position of the first correct match for each query:
\begin{equation}
    \text{MRR} = \frac{1}{|Q|} \sum_{i=1}^{|Q|} \frac{1}{\text{rank}_i}
    \label{eq:mrr}
\end{equation}
where $Q$ is the set of queries and $\text{rank}_i$ is the position of the first correct result for query~$i$. A high MRR indicates that relevant symbols consistently appear near the top of the ranked list.

All metrics are reported at the optimal threshold $\tau = 0.70$, identified via the sensitivity analysis described in Section~\ref{sec:threshold}.

\paragraph{Implementation details.}
The encoder is a ResNet-50 backbone pretrained on ImageNet,  producing 2048-D embeddings. The projection head maps to a 128-D space with temperature $\tau = 0.5$; label smoothing is set to $\varepsilon = 0.2$; the SupCon weight is $\lambda_s = 0.2$.
The class-balanced sampler uses $S = 4$ samples per class and a target batch size of $B = 192$, yielding $K = 32$ font classes per batch (128 font images) supplemented by $\sim$64 manuscript symbols. The two-stage training uses $T_w = 50$ warmup epochs; the model is then trained for up to 250 epochs total with early stopping (patience $= 25$). The style adaptation parameters are $k = 50$ neighbours and $\alpha = 0.7$.

\subsection{Threshold Sensitivity and Optimization.}
\label{sec:threshold}
A critical parameter in our retrieval pipeline is the similarity threshold $\tau \in [0, 1]$, which determines the minimum cosine similarity required to accept a match. A low threshold favours discovery increasing Cover at the cost of Precision, while a high threshold prioritises accuracy at the risk of missing valid alphabet members. To identify the optimal operating point, we conducted a systematic sensitivity analysis by varying $\tau$ from $0.50$ to $1.00$ in steps of $0.05$, measuring Precision@5 and Cover@5 across all fourteen test documents.

\begin{figure}[htbp]
    \centering
    \includegraphics[width=1\textwidth]{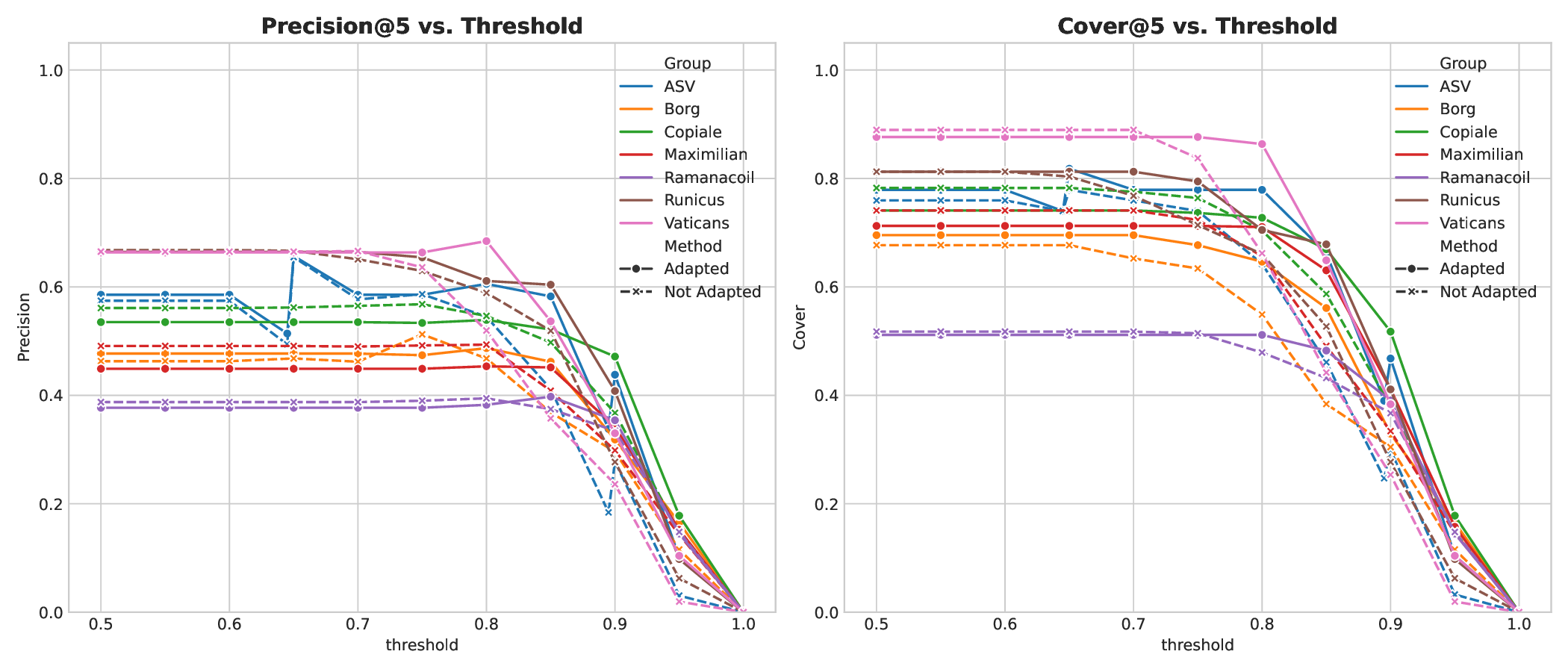}
    \caption{Sensitivity analysis showing the impact of the similarity threshold $\tau$ on Precision and Cover. Solid lines represent the proposed method with Style Adaptation; dashed lines represent the proposed method without Style Adaptation.}
    \label{fig:sensitivity}
\end{figure}

Figure~\ref{fig:sensitivity} reports the results aggregated by collection, with each line representing the average performance across the two pages of that collection. Solid lines correspond to our full method with style adaptation; dashed lines show the same model without adaptation. 
For $\tau \leq 0.70$, both Precision@5 and Cover@5 remain remarkably stable across all collections, indicating that the DANN-aligned embeddings produce 
well-separated clusters where correct matches score consistently above background noise. As $\tau$ increases beyond $0.70$, performance degrades sharply for the method without style adaptation, while the adapted variant maintains significantly 
higher scores, confirming that shifting the font query towards the local style manifold effectively rescues correct matches that would otherwise fall below a strict threshold. Based on these observations, we set $\tau = 0.70$ as the operating threshold for all subsequent experiments, as it maximises the trade-off 
between Cover and Precision while marking the point at which the benefit of style adaptation becomes most pronounced. All results in Tables~\ref{tab:comparison_results} and~\ref{tab:ablation_results} are reported at this threshold.

\subsection{Comparative Analysis and Ablation Study}
\label{sec:results}

\paragraph{Comparison with baseline systems.}
Table~\ref{tab:comparison_results} reports Precision@1, Precision@5, Cover@1, Cover@5, and MRR for our full method against eight baseline systems at $\tau = 0.70$.
We first consider \emph{zero-shot} baselines, which require no task-specific training. All zero-shot baselines (CLIP, DINOv2) are evaluated with frozen, publicly released weights and no fine-tuning on our font or manuscript data. Among these, CLIP ViT-L/14 achieves the strongest performance (P@1$= 0.401$, MRR$= 0.460$), outperforming both DINOv2 variants by a substantial margin. Notably, DINOv2-large underperforms DINOv2-base (P@1 $0.127$ vs.\ $0.222$), suggesting that larger patch-based Vision Transformers lose discriminative power when processing the small ($64\times64$) glyph crops typical of symbol spotting. This finding  is consistent with the known sensitivity of ViT architectures to input resolution. The strong relative performance of CLIP may be attributed to its contrastive pretraining on a vastly more diverse image distribution, which produces more general low-level features.

Task-specific training on our font and manuscript data improves results considerably. ResNet50-trained (P@1$= 0.457$) surpasses all zero-shot baselines, confirming that domain-specific supervision is necessary for this task and that no off-the-shelf foundation model can substitute for it. Interestingly, the untrained ResNet50-base (P@1$= 0.130$) performs \emph{worse} than its ResNet18 counterpart (P@1$= 0.160$), further evidence that raw model capacity does not help when the feature distribution is mismatched with the target domain.

Our full method (+SimCLR+DANN+adapt) achieves P@1$= 0.595$ and MRR$= 0.661$, outperforming the strongest baseline (CLIP ViT-L/14) by $+0.194$ in P@1 and $+0.201$ in MRR, a margin that leaves little ambiguity about the benefit of task-specific domain-adaptive training.

\begin{table}[htbp]
\centering
\caption{Comparison of our full method against zero-shot and task-specific baseline systems at $\tau = 0.70$. Zero-shot models require no training on the target domain; 
trained models are fine-tuned on our font and manuscript data.
Best result in \textbf{bold}, second best \underline{underlined}.}
\label{tab:comparison_results}
\begin{tabular}{lccccc}
\toprule
\textbf{Model}            & \textbf{P@1}    & \textbf{P@5}    & \textbf{Cover@1} & \textbf{Cover@5} & \textbf{MRR}    \\
\midrule
\textit{ResNet18-base}   & 0.1596          & 0.1524          & 0.1596           & 0.2909           & 0.2221          \\
\textit{ResNet50-base}   & 0.1302          & 0.1164          & 0.1302           & 0.2133           & 0.1678          \\
\textit{DINOv2-base}      & 0.2219          & 0.2050          & 0.2219           & 0.3166           & 0.2656          \\
\textit{DINOv2-large}     & 0.1269          & 0.1123          & 0.1269           & 0.1859           & 0.1530          \\
\textit{CLIP ViT-B/32}    & 0.3111          & 0.2620          & 0.3111           & 0.4367           & 0.3716          \\
\textit{CLIP ViT-L/14}    & 0.4012          & 0.3304          & 0.4012           & 0.5286           & 0.4604          \\
\textit{ResNet18-trained} & 0.3764          & 0.3617          & 0.3764           & 0.4064           & 0.3913          \\
\textit{ResNet50-trained} & \underline{0.4569} & \underline{0.4159} & \underline{0.4569} & \underline{0.5334} & \underline{0.4936}          \\
Our                       & \textbf{0.5949} & \textbf{0.5359} & \textbf{0.5949}  & \textbf{0.7326}  & \textbf{0.6609} \\
\bottomrule
\end{tabular}
\end{table}

\paragraph{Ablation study.}
Table~\ref{tab:ablation_results} isolates the contribution of each component by augmenting the ResNet50-trained baseline. Four configurations are evaluated: SimCLR alone, DANN alone, their combination, and the addition of style adaptation at retrieval time.

\begin{table}[b]
\centering
\caption{Ablation study at $\tau = 0.70$. Starting from the ResNet50-trained baseline, we progressively add SimCLR contrastive training (+SimCLR), domain adversarial training (+DANN), and embedding-space style adaptation (+adpt). The final row corresponds to our full method.
Best result in \textbf{bold}.}
\label{tab:ablation_results}
\begin{tabular}{lccccc}
\toprule
\textbf{Model}                   & \textbf{P@1}    & \textbf{P@5}    & \textbf{Cover@1} & \textbf{Cover@5} & \textbf{MRR}    \\
\midrule

\textit{Base (ResNet50-trained)} & 0.4569          & 0.4159          & 0.4569           & 0.5334           & 0.4936          \\
\textit{+SimCLR}                 & 0.4838          & 0.4535          & 0.4838           & 0.5736           & 0.5281          \\
\textit{+SimCLR+adpt}            & 0.5386          & 0.4908          & 0.5386           & 0.6596           & 0.5951          \\
\textit{+DANN}                   & 0.5246          & 0.4820          & 0.5246           & 0.6365           & 0.5766          \\
\textit{+DANN+adpt}              & 0.5451          & 0.4872          & 0.5451           & 0.6683           & 0.6044          \\
\textit{+SimCLR+DANN}            & 0.5856          & 0.5426          & 0.5856           & 0.7291           & 0.6525          \\
\textit{+SimCLR+DANN+adpt}       & \textbf{0.5949} & \textbf{0.5359} & \textbf{0.5949}  & \textbf{0.7326}  & \textbf{0.6609} \\
\bottomrule
\end{tabular}
\end{table}

Both SimCLR and DANN individually improve over the baseline (P@1 $0.484$ and $0.525$ respectively, vs.\ $0.457$), but their combination (+SimCLR+DANN, P@1$= 0.595$) yields a disproportionate gain that exceeds the sum of their individual contributions. This super-additive interaction suggests a productive complementarity: SimCLR shapes the geometry of the embedding space by clustering structurally similar symbols, while DANN aligns the font and manuscript distributions within that space. Each component creates conditions that make the other more effective.
Style adaptation provides a consistent further improvement across all configurations. Its contribution is largest when applied to SimCLR alone (+$0.055$ P@1) and smallest when applied to the full SimCLR+DANN model (+$0.009$ P@1). This diminishing return is expected: when the encoder has already reduced the domain gap via DANN, the residual shift that adaptation needs to correct is smaller. 
The fact that adaptation still helps even after DANN confirms that the two mechanisms address complementary aspects of the domain gap: global alignment (DANN) versus local, query-specific adjustment (style adaptation).
Taken together, these results establish that all three components make independent and complementary contributions to the final performance, with their combination achieving the best results across all evaluated metrics.

\subsection{Script Identification via Raw Cover}
\label{sec:script_id}

Beyond evaluating retrieval accuracy on known ground truth, we investigate whether the per-alphabet distribution of Raw-Cover@$k$ can serve as a \emph{script signature}, i.e. a fingerprint that identifies the underlying character families of an unknown document without any labelled examples. This use case is of direct relevance to paleographers and historians, who routinely face the problem of characterising an unfamiliar script before any transcription is available.

For this analysis we constructed a set of query alphabets consisting of the ten most frequent symbols from each of the following script families: Arabic, Copiale, Cuneiform, Digits, Esoteric, Greek, Latin, Runic, and Zodiac. Focusing on the most frequent symbols maximises the likelihood that the queried characters are actually present in the manuscripts, while keeping the query set compact. Cuneiform was included as a negative control: none of its symbols appear in any test document, so a well-calibrated system should produce near-zero Raw-Cover values for this alphabet across all collections. Raw-Cover@5 was computed at a high threshold $\tau = 0.95$ to minimise spurious matches and make the analysis as robust as possible; crucially, this metric requires no ground-truth labels and can therefore be applied in a fully unsupervised setting.

\begin{figure}[ht!]
  \centering
  \includegraphics[width=1\columnwidth]{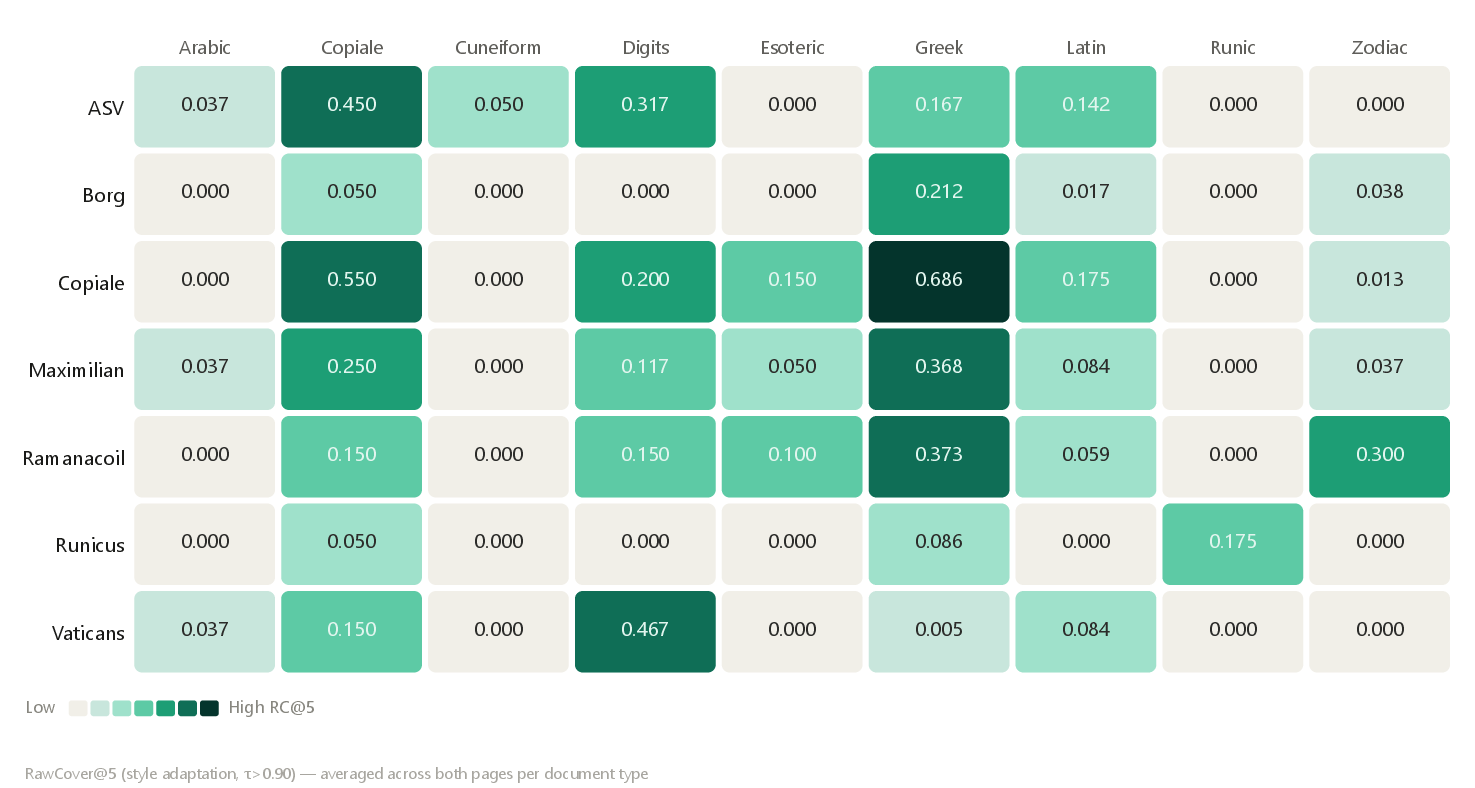}
  \caption{Raw-Cover@5 at $\tau = 0.95$, averaged across the two test pages of each collection (rows) for nine query alphabets (columns), computed without ground-truth labels.}
  \label{fig:heatmap_rc}
\end{figure}

\paragraph{Results.}
Figure~\ref{fig:heatmap_rc} reports Raw-Cover@5 averaged across the two pages of each collection. The heatmap reveals several interpretable patterns that align well with the known composition of each cipher alphabet.

ASV and Vaticans show high Raw-Cover for Digits ($0.317$ and $0.467$ respectively), consistent with the fact that both collections employ digit-based ciphers. Runicus, the only collection written in a runic script, is correctly identified by a high Raw-Cover for the Runic alphabet ($0.175$) and near-zero values for all others. The Copiale collection presents a particularly interesting case: it scores highest on the Greek alphabet ($0.686$), higher even than on its own Copiale alphabet ($0.550$). This is not an error; in the Copiale cipher, many symbols are indeed borrowed from the Greek script. This is a fact well known to expert in domain but here recovered automatically without any prior knowledge. 
Ramanacoil stands out for its high Zodiac coverage ($0.300$), reflecting the presence of several zodiac symbols in that collection.
The negative control behaves as expected: Cuneiform yields near-zero Raw-Cover throughout, with small residuals ($0.050$ for ASV) attributable to shape similarity with simple geometric primitives shared across scripts.

\begin{figure}[t!]
  \centering
  \includegraphics[width=0.9\columnwidth]{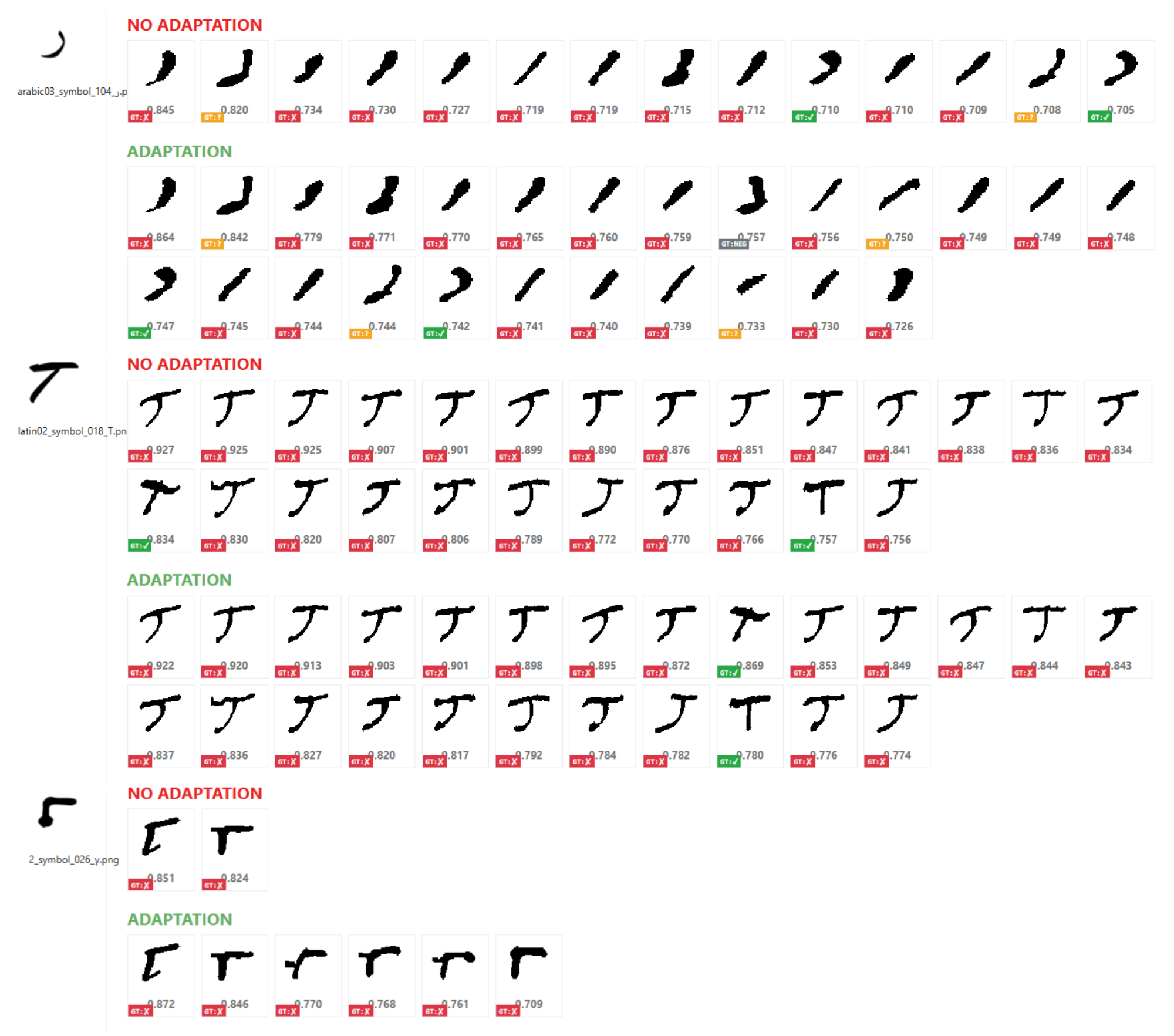}
  \caption{Each row shows a query font glyph (left) with its top retrieved manuscript symbols, with and without style adaptation. Red labels indicate incorrect matches; green correct ones. Despite high cosine similarity scores, retrieved symbols belong to different alphabet classes. 
  }
  \label{fig:false_positives}
\end{figure}

\paragraph{Error analysis.}
Two systematic sources of error limit the Raw-Cover values from reaching their theoretical maximum. 
First, query alphabets are built from each script's most frequent symbols in general, not guaranteed to all appear on the specific test pages (e.g. some Copiale digit symbols are simply absent).
Second, visually ambiguous symbol pairs (such as the Greek iota and the digit~1, or the Latin~O and the digit~0) produce cross-alphabet retrievals even at high thresholds. This ambiguity is intrinsic to the manuscripts themselves, where scribes frequently reused visually similar forms across different semantic functions, and represents a fundamental limit rather than a modelling failure.
Figure~\ref{fig:false_positives} illustrates representative examples of this phenomenon: despite high cosine similarity scores, the retrieved symbols belong to different alphabet classes, reflecting genuine visual ambiguity in the manuscripts rather than a modelling failure.\\
Taken together, these results demonstrate that Raw-Cover@$k$, computed in a fully unsupervised setting, provides a meaningful script-family fingerprint that is consistent with expert paleographic knowledge. 

\section{Conclusion}

We presented a zero-shot symbol spotting pipeline for encrypted historical manuscripts that requires no labelled target-document examples. A joint SimCLR+DANN encoder produces domain-invariant representations of font and manuscript glyphs, complemented by an embedding-space style adaptation that closes the residual domain gap at retrieval time without re-training. The ablation confirms all three components contribute independently and complementarily, with SimCLR+DANN (P@1=0.595) substantially outperforming either alone; against zero-shot foundation models, our method beats the strongest baseline (CLIP ViT-L/14) by +0.138 P@1, showing that large-scale pretraining alone cannot substitute for task-specific domain adaptation here.
Beyond retrieval accuracy, Raw-Cover@k, computed without any labels, provides a script-family fingerprint consistent with expert paleographic knowledge that can be considered as a practical tool for the initial exploration of undeciphered collections.

Two limitations remain. First, our experiments target detached-script collections, which simplify glyph segmentation; this is a dataset property rather than a method constraint, and future work will extend to cursive scripts with ligatures via alternative segmentation and synthetic augmentation. Second, and more fundamentally, the approach is query-driven: it can only detect symbols resembling some Unicode-renderable candidate alphabet, and offers no traction on scripts unrelated to any attested writing system, which would require an open-set or clustering-based formulation instead.
The longer-term goal is a fully automated pipeline that, given only a set of candidate alphabets and an unseen manuscript, can identify the underlying script family, providing a practical first step towards decipherment.

\section*{Acknowledgments}
This work was partially supported by Riksbankens Jubileumsfond, grant M24-0028 (DESCRYPT), the Spanish project PID2024-157778OB-I00 (SUKIDI), Ministerio de Ciencia e Innovación, the Departament de Cultura and the CERCA Program, Generalitat de Catalunya. A. Fornés acknowledges support from ICREA under the ICREA Academia programme (Departament de Recerca i Universitats, Generalitat de Catalunya).

%
%
%
\bibliographystyle{splncs04}
\bibliography{ref}

\end{document}